\def\year{2022}\relax
\documentclass[letterpaper]{article} 
\usepackage{aaai22}  
\usepackage{times}  
\usepackage{helvet}  
\usepackage{courier}  
\usepackage[hyphens]{url}  
\usepackage{graphicx} 
\usepackage{natbib}  
\usepackage{caption} 
\DeclareCaptionStyle{ruled}{labelfont=normalfont,labelsep=colon,strut=off} 
\usepackage{algorithm}
\usepackage{algorithmic}
\usepackage{amsfonts}
\usepackage{booktabs}
\usepackage{multirow}
\usepackage{bm}
\usepackage{amsmath}
\usepackage{xspace} 
\usepackage{color} 
\usepackage{subfigure}
\usepackage{multirow}
\usepackage{enumitem}

\newtheorem{definition}{Definition}

\newcommand{\B}[1]{{\bfseries #1}}
\newcommand{\model}{GeomGCL\xspace}
\newcommand{\gnn}{GeomMPNN\xspace}

\newcommand{\zhou}[1]{{\color{black}#1}}
\usepackage[normalem]{ulem}
\newcommand{\liadd}[1]{{\color{black}#1}} 
\newcommand{\hide}[1]{} 

\newcommand{\cat}{\ensuremath{\mathbin\Vert}}

\usepackage{newfloat}
\usepackage{listings}
\floatstyle{ruled}
\newfloat{listing}{tb}{lst}{}
\floatname{listing}{Listing}
\title{\model: Geometric Graph Contrastive Learning for \\ Molecular  Property Prediction}

\author{
    Shuangli Li,\textsuperscript{\rm 1, 2}\thanks{This work was done when the first author was an intern in Baidu Research under the supervision of the second author.}
    Jingbo Zhou,\textsuperscript{\rm 2}
    Tong Xu,\textsuperscript{\rm 1}
    Dejing Dou,\textsuperscript{\rm 2}
    Hui Xiong\textsuperscript{\rm 3}
    \\
}
\affiliations{
\textsuperscript{\rm 1}University of Science and Technology of China,  
\textsuperscript{\rm 2}Business Intelligence Lab, Baidu Research\\
\textsuperscript{\rm 3}Artificial Intelligence Thrust, The Hong Kong University of Science and Technology\\
lsl1997@mail.ustc.edu.cn,  \{zhoujingbo, doudejing\}@baidu.com, tongxu@ustc.edu.cn, xionghui@ust.hk
}

\begin{document}

\maketitle

\begin{abstract} \label{sec-abstract}

\zhou{\hide{In recent years}
Recently many efforts have been devoted to applying graph neural networks (GNNs) to molecular property prediction which is a fundamental task for computational drug and material discovery.  One of major obstacles to hinder the successful prediction of molecule property by GNNs is the scarcity of labeled data. Though graph contrastive learning (GCL) methods have achieved extraordinary performance with insufficient labeled data, most focused on designing data augmentation schemes for general graphs. However, the fundamental property of a molecule could be altered with the augmentation method (like random perturbation) on molecular graphs.  Whereas, the critical geometric information of molecules remains rarely explored under the current GNN and GCL architectures.
To this end, we propose a novel graph contrastive learning method utilizing the geometry of the molecule across 2D and 3D views, which is named \model. Specifically, we first devise a dual-view geometric message passing network (\gnn) to adaptively leverage the rich information of both 2D and 3D graphs of a molecule. The incorporation of geometric properties at different levels can greatly facilitate the molecular representation learning. Then a novel geometric graph contrastive scheme is designed to make both geometric views collaboratively supervise each other to improve the generalization ability of \gnn. We evaluate \model on various downstream property prediction tasks via a finetune process. Experimental results on seven real-life molecular datasets demonstrate the effectiveness of our proposed \model against state-of-the-art baselines.
}

\end{abstract}

\section{Introduction} \label{sec-introduction}



\zhou{
The prediction of molecular property has been widely considered as one of the most significant tasks in computational drug and material discovery \cite{goh2017deep,wu2018moleculenet,chen2018rise}.
Accurately predicting the property can help to evaluate and select the appropriate chemical molecules with desired characteristics for many downstream applications \hide{\cite{xiong2019pushing,yang2019analyzing,danel2020spatial,song2020communicative,shui2020heterogeneous,ijcai2021-309}}\cite{xiong2019pushing,yang2019analyzing,song2020communicative,ijcai2021-309}. 
%
}


\zhou{With the remarkable success of graph neural networks (GNNs) in various graph-related tasks in recent years \cite{wu2020comprehensive}, a number of efforts have been made from different directions to design GNN models for molecular property prediction like \cite{yang2019analyzing,danel2020spatial,maziarka2020molecule,song2020communicative,ijcai2021-309}. 
The fundamental idea is to regard the topology of atoms and bonds as a graph, and translate each molecule to a representation vector with powerful GNN encoders, followed by the prediction module for the specific property. 
}




\zhou{
Along the other line of development for GNNs, graph contrastive learning (GCL) methods \cite{wu2021self} have shown promising performance in many applications when there is a lack of sufficient labeled data. 
The scarcity of labeled data is \hide{just }one of the major obstacles to hinder the prediction performance of GNN models (\hide{and}as well as other deep learning models) for molecular property prediction. For example, it usually requires a high cost to collect the labeled data for certain computational drug discovery task. 
}

\zhou{\hide{There still needs a special research attention to apply}More special attention should be paid to developing GCL for molecular property prediction. Existing GCL methods usually adopt different data augmentation schemes for the graph\hide{data. However, such data augmentation method }, which may change the semantics of graphs across domains.\hide{ violate the fundamental chemical rules of molecules.} Most of the current GCL methods on molecular graphs are still based on such data augmentation paradigm which could inevitably alter the natural structure of a molecule. For example, \citet{you2020graph} proposes to drop atoms, perturb edges and mask attributes to augment the data. However, since each of atoms has an effect on the molecular property, such random dropping and perturbation of atoms could destroy the structure of a molecule. Though some other methods like MoCL \cite{sun2021mocl} adopt pre-defined sub-structures of the molecule to alleviate the problem of random corruption, such substitution rules still have a probability to violate the chemical principle. 
}

\begin{figure}[t]
\centering
\includegraphics[width=0.9\columnwidth]{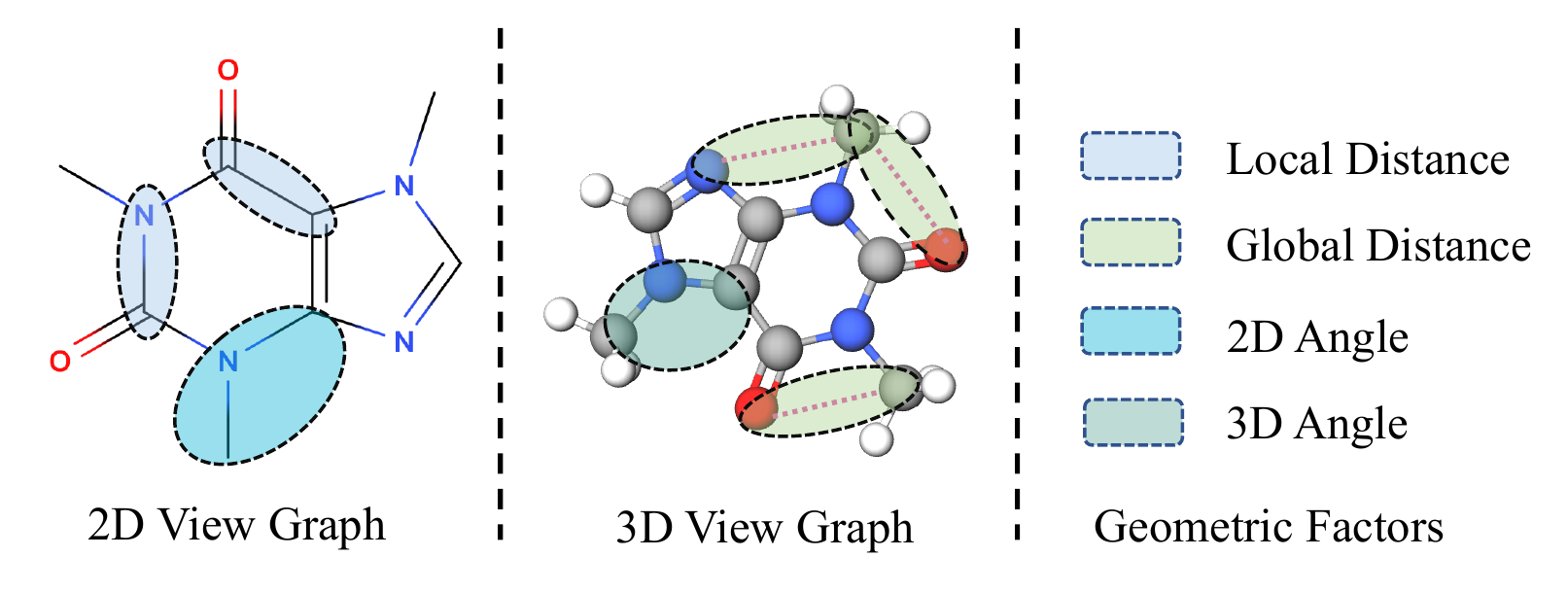}
\vspace{-2mm}
\caption{An illustrative example of geometric distances and angles in 2D view and 3D view graphs.}
\label{fig-dist-angle}
\vspace{-4mm}
\end{figure}

\zhou{
Our insight is to design a new GCL method for molecular property prediction from different geometric views without corrupting the molecular structure. As shown in Figure \ref{fig-dist-angle}, the molecules can be represented as two-dimensional (2D) and three-dimensional (3D) structural graphs. Although there are emerging geometric GNN models which can make use of multiple factors from the 2D chemical graph or 3D spatial graph \cite{maziarka2020molecule,klicpera_dimenet_2020,shui2020heterogeneous,danel2020spatial}, all of them capture the geometric information from a single view for representation learning. 
In practice, the 2D view and 3D view of a molecule are generated based on different methods: the 2D view derives directly from the structural formula of a chemical compound, while the 3D view is usually\hide{ has to be} estimated by conformation generation procedure \cite{hawkins2017conformation} using the tools like RDKit \cite{tosco2014bringing}.  
Considering that molecular graphs in 2D and 3D views can provide chemical and geometric information at different levels and further complement each other, there is a demand to develop a new paradigm for the molecule-driven contrastive learning without changing chemical semantics.
Therefore, such different views of one molecule provide a great opportunity for designing a unique graph contrastive learning scheme on molecular graphs.
}

%
%

To tackle the aforementioned challenges, we propose a novel geometric-enhanced graph contrastive learning model (\model) for molecular property prediction, which is equipped with the adaptive geometric message passing network (\gnn) as well as a contrastive learning strategy to augment the 2D-3D geometric structure learning process. Firstly, we devise the dual-channel geometric learning procedure to adapt the graph aggregation to both views with leveraging distance and angle information at different granularity levels. Secondly, we further aim to make both views \hide{enhance}complement each other for better geometric comprehension\hide{understanding}. Here, we take the molecular geometry into consideration and propose the 2D-3D geometric contrastive scheme to bridge the knowledge gap between geometric structure modeling and graph representation learning without labels. The representative\hide{beneficial} 3D spatial information can be distilled with the guidance of the stable 2D information, which provides the chemical semantics. The fusion of 2D and 3D graphs promotes the proposed \model extract more expressive representation for property prediction. To summarize, the main contributions of our work are as follows:

\begin{itemize}
    \item To the best of our knowledge, we are among the first to develop the contrastive learning method for molecular graphs  based on geometric views. By means of naturally contrasting the 2D and 3D view graphs of one molecule without any random augmentation process, our proposed \model makes \hide{the most}best of the consistent and realistic structures for better representation learning.
    \item The \model employs the dual-channel message passing neural network, which adaptively captures both 2D and 3D geometric information of molecular graphs. The additional spatial regularizer further preserves the relative relation of geometry and improves the performance.
    \item The experiments on seven molecular datasets demonstrate that the proposed model outperforms the state-of-the-art GNN and graph contrastive learning methods.
\end{itemize}






\section{Related Work} \label{sec-related}
The research topic of this paper is associated with \hide{the }molecular representation learning based on graph neural networks, especially the promising\hide{emerging} geometric and contrastive learning on molecular graphs. We will briefly discuss these topics.

\subsection{Molecular Representation Learning}
As the important basis of property prediction, molecular representation learning has been a popular research area. The earlier feature-based methods learn the fixed representations from molecular descriptors or chemical fingerprints \cite{rogers2010extended}, which ignore the graph structures and rely on the feature engineering. Recent years have witnessed the great advantage of graph neural networks (GNNs) in modeling graph data \cite{wu2020comprehensive}, much attention has been paid to applying GNN models to learn molecular graph representations. The graph convolution model \cite{duvenaud2015convolutional} is first introduced to encode molecular graphs based on atomic features. AttentiveFP \cite{xiong2019pushing} adopts the graph attention network to make the best of graph structures and has become one of the state-of-the-art methods for molecular property prediction. More recently, incorporating bond features into the message passing neural networks \cite{gilmer2017neural} turns into a trend of learning better representations. DMPNN \cite{yang2019analyzing} is proposed to perform the edge-based message passing process over the edge-oriented directed graph, which obtains atom and bond embeddings concurrently. The communicative message passing models \cite{song2020communicative,ijcai2021-309} further extend this work through improving the node-edge interactive kernels as well as applying the transformer framework to capture long-range dependencies. Nevertheless, these approaches can not deal with the geometry of molecules and lack the ability of learning the non-local correlations among atomic nodes on the molecular graph.


\subsection{Geometric Learning on Molecular Graphs}
In the field of deep learning, geometry-based methods have shown prominent performance \cite{bronstein2017geometric}. Since molecules have the geometric structures intrinsically, a few attempts have also been made to develop geometric graph learning models for the molecular graphs \cite{atz2021geometric}. From the 2D view of the molecular graph, MAT \cite{maziarka2020molecule} is designed to encode the inter-atomic distances with augmenting the attention mechanism in a transformer architecture. Meanwhile, there are some efforts to model the 3D molecular structures. SGCN \cite{danel2020spatial} simply utilizes the 3D coordinates to apply the aggregation process. Such an intuitive method is sensitive to the coordinate systems, which leads to the poor performance of learning geometric graphs. Furthermore, several models that are invariant to translation and rotation
of atom coordinates are proposed through designing the geometric kernels \cite{klicpera_dimenet_2020} or strengthening the node-edge interactions with geometric information \cite{shui2020heterogeneous}. However, these studies have shown not just strengths but also some limitations. Firstly, most of the efficient molecule-oriented geometric learning methods target at the quite small molecules for quantum property prediction. Secondly, none of these methods incorporate 2D and 3D geometric information simultaneously. To overcome these limitations, we propose to learn the 2D-3D geometric factors adaptively and synergistically.

\subsection{Contrastive Learning on Molecular Graphs}
Along the other line of development, graph contrastive learning methods \cite{wu2021self} have their own advantages and have achieved extraordinary performance in many applications \cite{you2020graph,qiu2020gcc,wang2021self}. However, the existing models designed for molecular graphs receive little attention. InfoGraph \cite{sun2019infograph} manages to maximize the mutual information between the representations of the graph and its substructures to guide the molecular representation learning. To alleviate the problem of random corruption on molecular graphs which may alter the chemical semantics, MoCL \cite{sun2021mocl} adopts the domain knowledge-driven contrastive learning framework at both local- and global-level to preserve the semantics of graphs in the augmentation process. However, the learning ability of such model depends on the well-designed substitution rules. The deficiency of geometric information in graph contrastive learning limits the capability of effective molecular representation learning. To this end, we develop a novel contrastive learning model to integrate the geometry of molecules with chemical semantics by means of contrasting the 2D and 3D view graphs.















\section{Preliminaries}\label{sec-pre}
\hide{
In this section, we first formally define the problem of molecular graph representation learning with \hide{downstream prediction tasks}geometric information injection, and then detail the geometric graph construction process for both 2D and 3D molecular views.

\subsection{Problem Definition}}
Generally, a molecular graph with geometric information can be represented as $\mathcal{G}=(\mathcal{V},\mathcal{E}, \bm{C})$, where $\mathcal{V}$ and $\mathcal{E}$ denote the node (atom) set and edge (bond) set respectively. $\bm{C} \in \mathbb{R}^{|V| \times d}$ denotes the coordinate matrix for atoms, where $d \in \{2,3\}$ is the spatial dimension.
Given a molecule, the specific 2D and 3D view graphs are defined as following. 

\begin{definition}[2D View Graph.] The 2D view graph is defined as $\mathcal{G}^{2d}=(\mathcal{V},\mathcal{E}^{2d}, \bm{C}^{2d})$, where the edges in $\mathcal{E}^{2d}$ correspond to the primary covalent bonds in the molecule, and $\bm{C}^{2d}_v = \{x,y\}$ denotes the coordinate of atom $v$.
\end{definition}

\begin{definition}[3D View Graph.]
Similarly, the 3D view graph can be represented as $\mathcal{G}^{3d}=(\mathcal{V},\mathcal{E}^{3d}, \bm{C}^{3d})$. Note that the generated coordinate in three dimensional space is non-deterministic by means of the estimation algorithm, thus we repeatedly generate coordinates $P$ times and $\bm{C}^{3d}=\frac{1}{P}\sum^{P}_{p=1}\bm{C}^{3d,p}$. The edge set $\mathcal{E}^{3d}$ is constructed based on the 3D spatial coordinates, which contains all edges whose distances are smaller than the cutoff threshold $d_{\theta}$. It can be formulated as $\mathcal{E}^{3d}=\{e_{uv}|dist(\bm{C}^{3d}_u,\bm{C}^{3d}_v)<d_{\theta}\}$.
\end{definition}

\B{Problem Statement.} Given a molecule, we can construct the 2D view graph $\mathcal{G}^{2d}$ and 3D view graph $\mathcal{G}^{3d}$. Let $\bm{X}$ be the atom feature matrix and $\bm{E}$ be the bond feature matrix, our goal is to train a geometric graph encoder $f(\mathcal{G}^{2d},\mathcal{G}^{3d},\bm{X},\bm{E})$ to learn the molecular representation vector $\bm{h}$ without any label information. 
Then the well-trained model is utilized for various downstream property prediction tasks through the finetune process.

\hide{\subsection{Molecular Graph Construction}}
\section{Model Framework}\label{sec-model}
\begin{figure*}[t]
\centering
\includegraphics[width=0.9\textwidth]{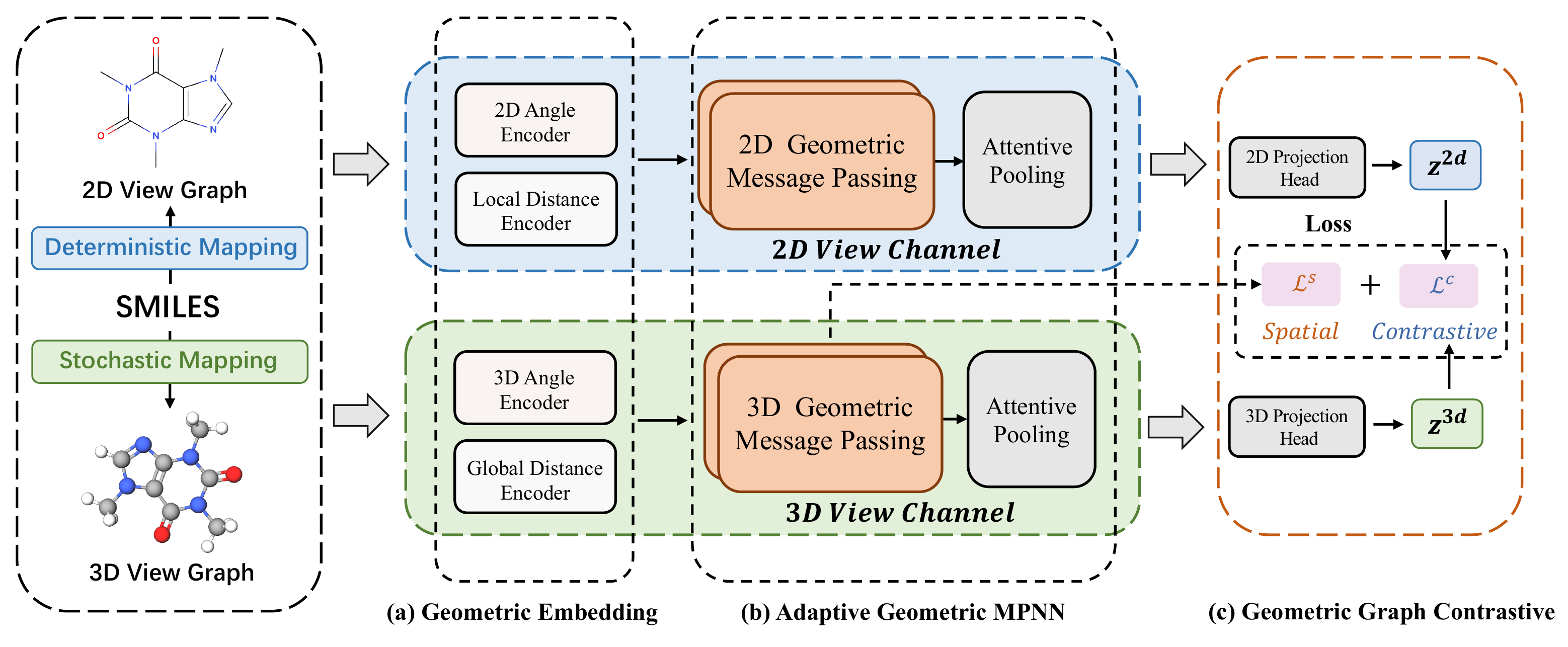} %
\vspace{-3mm}
\caption{Illustration of geometric graph contrastive learning framework (\model) on molecular graphs.}
\label{fig-model}
\vspace{-4mm}
\end{figure*}

In this section, we present the proposed contrastive learning framework \model with leveraging dual geometric views of the molecular graph.
As shown in Figure \ref{fig-model}, after the derivation of 2D and 3D view graphs from the original molecule in SMILES format, \model equips with a dual-channel geometric message passing architecture (\gnn) to learn the representations of both graphs adaptively. By contrast with the deterministic graph structure of 2D view, the 3D structure of molecule is always \hide{approximated}calculated through a stochastic process of 3D conformation generation.
\hide{which means that such uncertain geometric information is not always beneficial for molecular representation learning.} Consequently, while the 3D view graph\hide{with 3D structure} contains more abundant geometric structure, such uncertain information is not always beneficial for molecular representation learning. To this end, we propose to bypass this challenge by adopting a geometric contrastive learning strategy across 2D and 3D views. This geometric-view supervision mutually makes both views complement each other\hide{NEED TO POLISH}. On the one hand, \gnn can distill the valuable 3D structure under the guidance of 2D view. On the other hand, it helps to inject 3D geometric information for better 2D molecular representation learning. In the following sections, we use the bold letters (\hide{e.g., }$\bm{l}$, $\bm{r}$, $\bm{\phi}$, $\bm{\theta}$, $\bm{e}_{uv}$ or $\bm{a}_u$) to represent embeddings of the corresponding symbolic indicators (\hide{e.g., }$l$, $r$, $\phi$, $\theta$, $e_{uv}$ or $a_u$).



\subsection{Geometric Embedding}

Since the primary 2D or 3D coordinates are changeable and inconsistent across different coordinate systems, we manage to calculate the definite geometric factors (i.e., angle and distance) and then utilize radial basis functions (RBF) to obtain dual-level geometric embeddings. As illustrated in Figure \ref{fig-dist-angle}, the local distance $l$ and 2D angle $\phi$ refer to the distance and angle \hide{of}based on the covalent bonds respectively, which carry the critical chemical information. In the 3D view graph, the global distance $r$ further provides non-local correlations in a molecule, while the 3D angle $\theta$ indicates the spatial distribution of global connections. Following the previous work \cite{shui2020heterogeneous}, we adopt several RBF layers to encode diverse geometric factors:
\begin{align}
\label{eq-rbf}
    \bm{l}=RBF(l) &= \underset{k=1}{\overset{K}{\frown}}{\rm exp}\big (-\beta_{l}({\rm exp}(-l)-\mu_{l,k})^2 \big)
    \\
    \bm{r}=RBF(r) &= \underset{k=1}{\overset{K}{\frown}}{\rm exp}\big (-\beta_{r}({\rm exp}(-r)-\mu_{r,k})^2 \big)
    \\
    \bm{\phi}=RBF(\phi) &= \underset{k=1}{\overset{K}{\frown}}{\rm exp}\big (-\beta_{\phi}(\phi-\mu_{\phi,k})^2 \big)
    \\
    \bm{\theta}=RBF(\theta) &= \underset{k=1}{\overset{K}{\frown}}{\rm exp}\big (-\beta_{\theta}(\theta-\mu_{\theta,k})^2 \big)
\end{align}
where $\frown$ is the concatenation operation over scalar values to form a K-dimensional geometric embedding. For local and global distances, the K central points $\{\mu_{*,k}\}$ are uniformly selected between ${\rm exp}(-*)$ ($*$ is $l$ or $d$) and 1, while $\beta_{*}=(\frac{2}{K}(1-{\rm exp}(-*))^{-2}$. For 2D and 3D angles, each $\mu_{*,k}$ is between 0 and $\pi$ with $\beta_{*}=(\frac{2\pi}{K})^{-2}$, where $*$ denotes $\phi$ or $\theta$. 

\subsection{Adaptive Geometric Message Passing}
\begin{figure}[t]
\centering
\includegraphics[width=1.0\columnwidth]{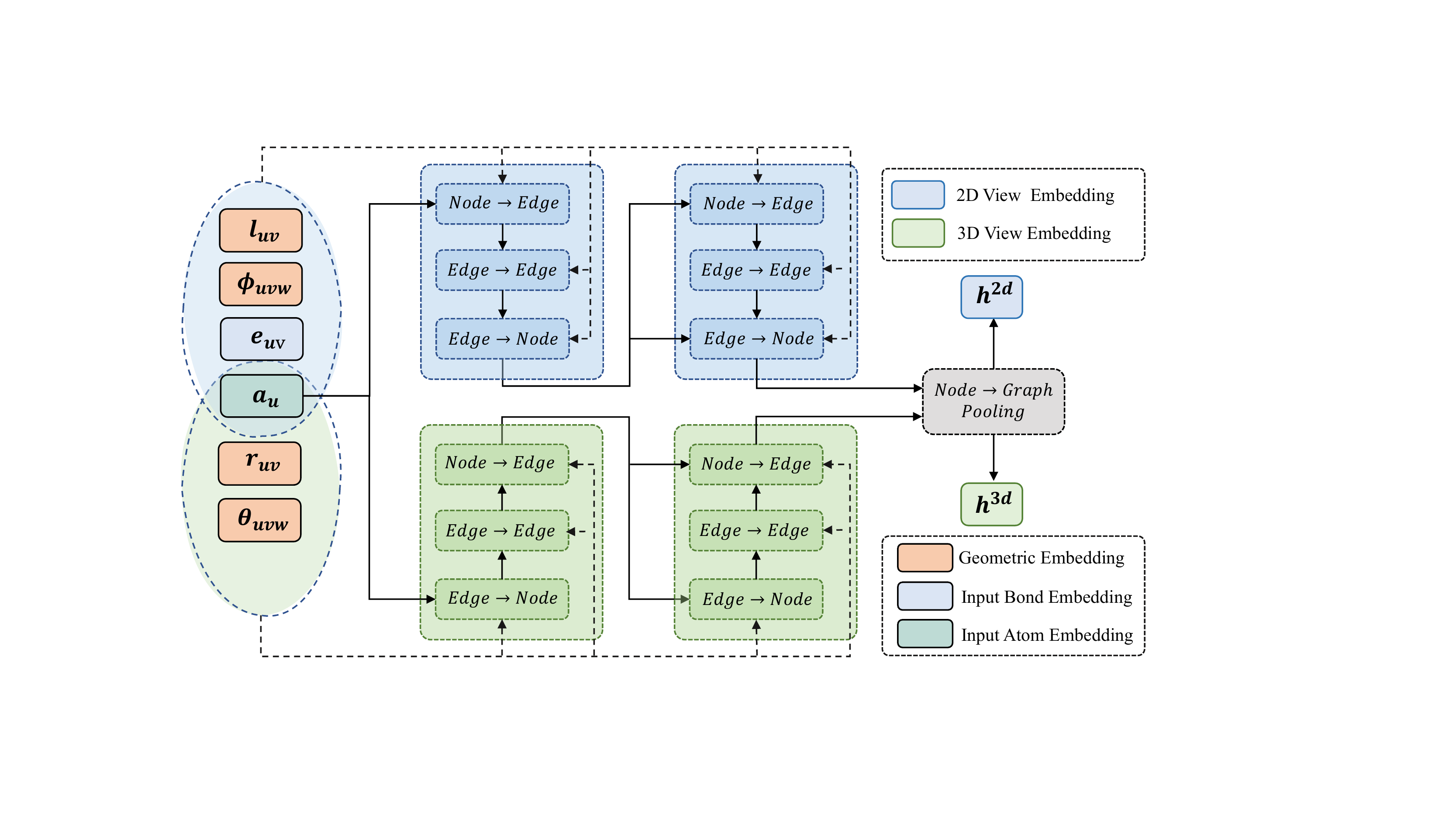}
\vspace{-6mm}
\caption{The architecture of \gnn layers.}
\label{fig-mpnn}
\vspace{-4mm}
\end{figure}

As shown in Figure \ref{fig-mpnn}, we further design an adaptive message passing scheme (\gnn) to learn the topological structures of molecules with geometric information in a Node-Edge interactive manner. On the whole, \gnn consists of \textit{Node$\bm{\rightarrow}$Edge}, \textit{Edge$\bm{\rightarrow}$Edge} and \textit{Edge$\bm{\rightarrow}$Node} three-stage message passing layers to iteratively update the node and edge embeddings, followed by a \textit{Node$\bm{\rightarrow}$Graph} attentive pooling process. Both of the dual-channel networks generally follow such architecture and can adaptively learn the 2D and 3D geometric factors with fine-grained designs.
\subsubsection{(i) Node$\bm{\rightarrow}$Edge Message Passing.}
Since only the existing bonds in a molecule have the initial edge features (i.e., bond features), the edge embedding should be firstly updated through aggregating the pairwise node embeddings with involving the associated features. To enrich the connection information from different aspects, we use MLP to integrate the chemical bond feature $\bm{e}^{0}_{uv}$ and global distance embeddings $\bm{r}_{uv}$ for 2D- and 3D-edge embeddings respectively: 
\begin{align}
\label{eq-n2e}
    \bm{e}^{2d,t}_{uv}&=MLP(\bm{a}^{2d,t-1}_u\cat\bm{a}^{2d,t-1}_v\cat\bm{e}^{0}_{uv})
    \\
    \bm{e}^{3d,t}_{uv}&=MLP(\bm{a}^{3d,t-1}_u\cat\bm{a}^{3d,t-1}_v\cat\bm{r}_{uv})
\end{align}
where $\bm{e}^{2d,t}_{uv}$ and $\bm{e}^{3d,t}_{uv}$ are edge embeddings at $t$-th layer, $\bm{a}_u$ is the node (atom) embedding, the superscript $2d$ or $3d$ indicates the view channel, and $\cat$ represents the concatenation operator. Then the dual-channel edge embeddings can contain both geometric and chemical semantic information.


\subsubsection{(ii) Edge$\bm{\rightarrow}$Edge Message Passing.}
Different from the general graph, both the 2D view and 3D view graphs of a molecule have the unique geometric attributes, which can significantly influence the specific property of the molecule\hide{cite?}. After the derivation of edge embeddings, \gnn performs an edge$\bm{\rightarrow}$edge message passing process to perceive the geometric\hide{or spatial?} distribution in the molecule through 2D and 3D angle-aware aggregations. Considering that the neighbors in 2D view are more sparse by contrast with the neighbors in 3D view, we develop the well-directed layers for both views. Specifically, for 2D view graph learning, the following angle-injected function is employed to update the 2D edge embedding:
\begin{equation}
    \bm{e}^{2d,t}_{uv}=\sum_{e_{wu} \in \mathcal{A}(e_{uv})} W^t_{\phi}\bm{\phi}_{wuv} \odot (W_{e}^{t} \bm{e}^{2d,t}_{uv})
\end{equation}
where $\mathcal{A}(e_{uv})$ denotes the set of neighboring edges of the edge $e_{uv}$, $\bm{\phi}_{wuv}$ is the 2D angle embedding between the edge $e_{wu}$ and $e_{uv}$, $\odot$ is the element-wise dot operation, $W^t_{\phi}$ and  $W_{e}^{2d,t}$ are learnable parameters. For the 3D view graph, there are sufficient neighbors around each edge. Inspired by the recent work \cite{li2021structure}, we divide the neighboring edges of each target edge into several angle domains $\mathcal{A}_1, ..., \mathcal{A}_n$ according to 3D spatial angle $\theta$. Then we apply a hierarchical aggregation process among edges for 3D view graph learning, which consists of local and global stages:
\begin{align}
\label{eq-e2e-3d}
    \bm{e}^{3d,t}_{uv,i}&=\sum_{e_{wu} \in \mathcal{A}_i(e_{uv})} W^t_{\theta,i}\bm{\theta}_{wuv} \odot (W_{e,i}^t \bm{e}^{3d,t}_{uv})
    \\
     \bm{e}^{3d,t}_{uv}&=\operatornamewithlimits{||}_{i=1}^n {\rm Pool} \big (\{\bm{e}^{3d,t}_{uv,i}|1 \le i\le n\}\big ) \odot \bm{e}^{3d,t}_{uv,i}
\end{align}
where $\bm{e}^{3d,t}_{uv,i}$ is the aggregated 3D edge embedding at $i$-th angle domain through the local stage, $W_{e,i}$ and $W^t_{\theta,i}$ are trainable parameters, $\bm{\theta}_{wuv}$ is the 3D angle embedding between the edge $e_{wu}$ and $e_{uv}$. ${\rm Pool}$ means the max pooling function over all $n$ local edge embeddings, which can generally extract the high-level spatial distribution information to strengthen the geometric structure learning.
\subsubsection{(iii) Edge$\bm{\rightarrow}$Node Message Passing.}
After obtaining the angle-aware edge embeddings $\bm{e}^{2d,t}_{uv}$ and $\bm{e}^{3d,t}_{uv}$, we apply the edge$\bm{\rightarrow}$node message passing to fulfil the propagation process from the edge back to the node. The essential distance factor between nodes is well-considered via the similar adaptive scheme with the previous edge$\bm{\rightarrow}$edge component:
\begin{align}
\label{eq-e2a}
     \bm{a}^{2d,t}_{v}&=\sum_{e_{uv} \in \mathcal{D}(a_{v})} W^t_{l}\bm{l}_{uv} \odot (W_{a}^{t} \bm{e}^{2d,t}_{uv})
    \\
     \bm{a}^{3d,t}_{v}&=\operatornamewithlimits{||}_{i=1}^m \sum_{e_{uv} \in \mathcal{D}_i(a_{v})} W^t_{r,i}\bm{r}_{uv} \odot (W_{a,i}^t \bm{e}^{3d,t}_{uv})
\end{align}
where $W^t_{l}$, $W_{a}^{t}$, $W^t_{r,i}$ and $W_{a,i}^t$ are learnable parameters, $\bm{l}_{uv}$ and $\bm{r}_{uv}$ are the geometric embeddings of local and global distances respectively, $\mathcal{D}(a_u)$ is the set of all neighboring edges for node $a_u$ in 2D view, $\mathcal{D}_i(a_u)$ is the set of neighboring edges located in $i$-th distance domain among all $m$ divided domains in 3D view. \hide{Since all geometric factors thoroughly enrich the node and edge embeddings in both 2D and 3D views via the above message passing layers, the final representations $\bm{a}^{2d,T}_{v}$ and $\bm{a}^{3d,T}_{v}$ can reflect the graph topology with molecular geometry.} 

\subsubsection{(iv) Node$\bm{\rightarrow}$Graph Attentive Pooling.}
Since all geometric factors thoroughly enrich the node and edge embeddings in both 2D and 3D views via \hide{the above}$T$ stacked message passing layers, the final representations $\bm{a}^{2d,T}_{v}$ and $\bm{a}^{3d,T}_{v}$ can reflect the graph topology along with the molecular geometry. To further get the graph-level representation with identifying the important nodes, we follow \cite{xiong2019pushing} and adopt the node$\bm{\rightarrow}$graph attentive pooling layer. For simplicity, we use $\bm{a}_v$ to represent $\bm{a}^{2d,T}_{v}$ or $\bm{a}^{3d,T}_{v}$ and use $\bm{h}$ to represent $\bm{h}^{2d}$ or $\bm{h}^{3d}$. The graph-level embedding $\bm{h}$ is updated iteratively through the attentive propagation process, which starts with the initial embedding $\bm{h}^{0}=\sum_v\bm{a}_v$.
\begin{align}
\label{eq-pool}
     \bm{g}^{t}&=\sum_{a_{v} \in \mathcal{V}} {\rm softmax}(\bm{h}^{t}\cat \bm{a}_v) W_g^{t}\bm{a}_v
     \\
     \bm{h}^{t+1}&={\rm GRU}(\bm{h}^{t},\bm{g}^{t}), \  t=0,1,..,T_g
\end{align}
where $\bm{g}^t$ is the global context message at $t$-th pooling layer, which aggregates the valuable nodes from the full set $\mathcal{V}$. After performing $T_g$ \hide{attentive }pooling layers, the final graph representations $\bm{h}^{2d}$ for 2D view and $\bm{h}^{3d}$ for 3D view are acquired. 

\subsection{Geometric Contrastive Optimization}
Despite the great progress made in the study of graph contrastive learning, the distinctive semantics and geometry\hide{geometrics} of molecular graphs are always ignored. As a result, the conventional data augmentations may change the targeted molecular property as stated in \cite{sun2021mocl}, which emphasized the importance of domain knowledge to generate molecule variants. To tackle this challenge, we intend to enhance the molecular representation learning from the perspective of geometric contrast. As shown in Figure \ref{fig-model}(c), the correlated 2D and 3D views can supervise each other without constructing additional fake\hide{augmented} samples. First, the 2D and 3D projection heads are adopted to map the representations of two views into the space for contrastive learning:
\begin{align}
\label{eq-proj-head}
     \bm{z}^{2d}_i=MLP(\bm{h}^{2d}_i), \ \ \bm{z}^{3d}_i=MLP(\bm{h}^{3d}_i)
\end{align}
Our goal is to make the maximum consistency between 2D-3D positive pairs $\{\bm{z}^{2d}_i,\bm{z}^{3d}_i\}$ \hide{of the same molecule }compared with negative pairs. Given one batch with $N$ molecules, we have the following contrastive loss function under 2D-3D geometric views:
\begin{align}
\label{eq-contrastive-loss}
     \mathcal{L}^c_i &= \mathcal{L}^{2d,c}_i + \mathcal{L}^{3d,c}_i \nonumber \\
                   &= {\rm -log}\frac{e^{\left \langle\bm{z}^{2d}_i,\bm{z}^{3d}_i\right \rangle/\tau}}{\sum_{j=1}^{N}e^{\left \langle\bm{z}^{2d}_i,\bm{z}^{3d}_j\right \rangle/\tau}}{\rm -log}\frac{e^{\left \langle\bm{z}^{3d}_i,\bm{z}^{2d}_i\right \rangle/\tau}}{\sum_{j=1}^{N}e^{\left \langle\bm{z}^{3d}_i,\bm{z}^{2d}_j\right \rangle/\tau}}
\end{align}
where $\left \langle\cdot\right \rangle$ denotes the inner product to measure the similarity, $\tau$ is the scale parameter. Since $\bm{z}^{2d}_i$ and $\bm{z}^{3d}_i$ are two embeddings in different views from the same molecule, they are regarded as a positive pair while the remaining pairs in the batch are considered as negative pairs. Furthermore, to reflect the local spatial correlations across 3D geometric domains, we propose additional constraints as a spatial regularization technique. The key idea of this regularizer is to encourage the transformation matrices of adjacent angle domains to be similar to each other:
\begin{equation}
    \mathcal{L}^s = \sum_{t=1}^{T}\sum_{i=1}^{n-1}\cat W^t_{\theta,i+1}-W^t_{\theta,i} \cat^2
\end{equation}
Finally, we combine the spatial and contrastive loss and arrive at the following objective function:
\begin{equation}
    \mathcal{L}_{gcl} = \sum\nolimits_{i=1}^{M}\mathcal{L}^c_i + \lambda\mathcal{L}^s
\end{equation}
where $M$ is the number of molecules in the dataset, $\lambda$ is the trade-off parameter that controls the importance of spatial regularizer to better guide the representation learning.

\subsection{Downstream Inference}
When \model has been optimized through the geometric contrastive learning process, we utilize the well-trained 2D geometric MPNN $f_{2d}(\cdot)$ and 3D geometric MPNN $f_{3d}(\cdot)$ for downstream applications. The representations of two views are combined to predict the molecular properties via the finetune process. Formally, the prediction head can be written as follows:
\begin{equation}
    \hat{y} = MLP\Big(MLP\big(f_{2d}(\mathcal{G}^{2d})\big) + MLP\big(f_{3d}(\mathcal{G}^{3d})\big)\Big)
\end{equation}
For different tasks, we use the cross entropy loss function $\mathcal{L}_{ce}$ for classification loss $\mathcal{L}_{cls}$ and use the L1 loss function $\mathcal{L}_{1}$ for regression loss $\mathcal{L}_{reg}$. The spatial regularizer is also adopted for better performance.
\begin{align}
    \mathcal{L}_{cls} &= \mathcal{L}_{ce}(\hat{y},y) + \lambda\mathcal{L}^s \\
    \mathcal{L}_{reg} &= \mathcal{L}_1(\hat{y},y) + \lambda\mathcal{L}^s 
\end{align}
where $\hat{y}$ denotes the predicted value and $y$ is the measured true value of one specific molecular property.

\section{Experiments}
In this section, we conduct experiments on seven well-known benchmark datasets to demonstrate the effectiveness of\hide{ our proposed} \model for molecular property prediction.

\subsection{Experiment Settings}
\subsubsection{Datasets.} To evaluate the performance of our proposed model with the existing molecular representation learning methods, we use seven molecular datasets from MoleculeNet \cite{wu2018moleculenet} including ClinTox, Sider, Tox21 and ToxCast four physiology datasets for graph classification tasks, as well as ESOL, FreeSolv and Lipophilicity three physical chemistry datasets for graph regression tasks. The main statistics of datasets are summarized in Table \ref{table-dataset}.

\subsubsection{Baselines.}
We compare the proposed \model with a variety of state-of-the-art baseline models for molecular property prediction, which includes molecular message passing-based methods, geometry learning-based GNN methods, and graph contrastive learning methods. The first group consists of three well-designed message passing neural networks. AttentiveFP \cite{xiong2019pushing} adopts the graph attention network for molecular representation learning. DMPNN \cite{yang2019analyzing}\hide{, CMPNN \cite{song2020communicative}} and CoMPT \cite{ijcai2021-309} are message passing models with considering edge features in a node-edge interactive manner. Geometry learning-based models contain several GNN approaches. SGCN \cite{danel2020spatial} directly encodes the atomic position information in the aggregation process. MAT \cite{maziarka2020molecule} incorporates the local geometric distance into the graph-based transformer model. To comprehensively reflect the superiority of our proposed model, we also compare \model against with HMGNN \cite{shui2020heterogeneous} and DimeNet \cite{klicpera_dimenet_2020}, both of which can learn the geometric distance and angle factors in 3D space for quantum property prediction. Besides, the recent graph contrastive models are compared to show the power of our proposed 2D-3D geometric contrastive learning strategy. InfoGraph \cite{sun2019infograph} maximizes the mutual information between nodes and graphs, while MoCL \cite{sun2021mocl} introduces the multi-level domain knowledge for molecular graphs in a well-designed contrastive learning framework.

\begin{table}
	\centering
	\begin{tabular}{ccccccc}
		\toprule
		Dataset	& \# Tasks	& Task Type	& \# Molecules \\
		\midrule
		ClinTox	& 2  & Classification	& 1484	\\
		Sider	& 27 & Classification	& 1427	\\
		Tox21	& 12 & Classification	& 7831	\\
		ToxCast	& 617 & Classification	& 8597	\\
		ESOL	& 1 & Regression	& 1128	\\
		FreeSolv	& 1 & Regression	& 643	\\
		Lipophilicity	& 1 & Regression	& 4200	\\
		\bottomrule
	\end{tabular}
		\caption{Statistics of seven molecular datasets.}
	\vspace{-3mm}
	\label{table-dataset}
\end{table}
\renewcommand\arraystretch{1.0}

\begin{table*}[t]
    \centering
	\scalebox{1.}{\begin{tabular}{c|c|c|c|c|c|c|c|c|c}
		\toprule
		\multirow{2}{*}{Model} & \multicolumn{5}{c|}{Graph Classification (ROC-AUC) $\uparrow$} & \multicolumn{4}{c}{Graph Regression (RMSE) $\downarrow$}   \\
		\cline{2-10}
		\rule{0pt}{10pt}
			&	ClinTox	&	Sider	&	Tox21	& ToxCast & Cls.Ave &	ESOL	&	FreeSolv	&	Lipophilicity & Reg.Ave	\\
		\midrule
		AttentiveFP  & 0.808 & 0.605  & 0.835  & 0.743  & 0.748  &	0.578  & 1.034 & 0.602 & 0.738   \\
        DMPNN  & 0.886 & 0.637  & 0.848  & 0.743  & 0.779  &	0.647  & 1.092 & 0.591 & 0.777   \\
        \hide{CMPNN  & 0.907 & 0.643  & 0.849  & 0.760  & 0.789  &	0.574  & 0.924 & 0.562 & 0.687   \\}
        CoMPT  & 0.877 & 0.626  & 0.836  & 0.755  & 0.774  &	0.589  & 1.103 & 0.590 & 0.761   \\
        \midrule
		SGCN  & 0.825 & 0.560  & 0.769  & 0.656  & 0.703  &	1.329  & 2.061 & 1.075 & 1.488   \\
        MAT  & 0.898 & 0.619  & 0.834  & 0.735  & 0.772  &	0.624  & 1.059 & 0.705 & 0.796   \\
        HMGNN  & 0.680 & 0.607  & 0.794  & 0.702  & 0.696  & 0.701  & 1.207 & 0.720 & 0.876   \\
        DimeNet  & 0.760 & 0.615 & 0.780  & 0.645  & 0.7000  &	0.633  & 0.978 & 0.614 & 0.742   \\
		\midrule
        InfoGraph  & 0.781 & 0.585  & 0.793  & 0.705 & 0.716  & 0.914  & 2.104 & 0.845 & 1.288   \\
		MoCL  & 0.739 & 0.629  & 0.824  & 0.718  & 0.727  &	0.934  & 1.478 & 0.742 & 1.051   \\
		\midrule
		\gnn  & 0.900 & 0.638  & 0.838  & 0.743  & 0.780  &	\B{0.555}  & 0.913 & 0.578 & 0.682   \\
        \model    &	\B{0.919} & \B{0.648} & \B{0.850} & \B{0.763 } &	\B{0.796} &	0.575  & \B{0.866}  & \B{0.541}  & \B{0.661}  \\
		\bottomrule
	\end{tabular}}
		\caption{Experimental results of our proposed \gnn and \model along with all baselines on seven molecular graph datasets. Cls.Ave and Reg.Ave denote the average results of classification and regression tasks, respectively.
	\vspace{-1mm}
	}

	\label{table-expriemnt}
	\vspace{-2mm}
\end{table*}

\subsubsection{Implementation Details.}
Following the previous works, we evaluate all methods through the k-fold cross-validation experiments, and we set k as 10 to \hide{sufficiently} report the robustly average experimental results. As recommended by the MoleculeNet benchmarks \cite{wu2018moleculenet}, we randomly split each dataset into training, validation, and testing set with a ratio of 0.8/0.1/0.1. The validation set is used for early stop and model selection. \hide{We use ROC-AUC metric for graph classification tasks and RMSE metric for graph regression tasks.}We use ROC-AUC and RMSE metrics for graph classification and graph regression tasks respectively.

The 3D structures of molecules are generated for $P=50$ times through the stochastic optimization algorithm of Merck Molecular Force Field (MMFF), which is implemented in the RDKit package \cite{tosco2014bringing}. For our model,  We use Adam optimizer for model training with a learning rate of 1e-3. We set the batch size as 256 for contrastive learning and 32 for finetuning with the scale parameter $\tau=0.5$. The hidden size of all models is set to 128. The cutoff distance $d_{\theta}$ is determined (4 Å or 5 Å) according to the size of the molecule on each dataset. We set the dimension $K$ of geometric embedding as 64. The numbers of 3D angle domains and global distance domains are set to 4. The balancing hyper-parameter $\lambda$ is set to 0.01 according to the performance on validation set. For baseline models, we tune parameters of each method based on recommended settings in the paper to ensure the best performance. As a general setting \cite{sun2019infograph,sun2021mocl}, our proposed \model and contrastive learning baselines are pretrained on molecular graphs of each dataset, and then we finetune the model for the downstream task on the same dataset.

\subsection{Performance Evaluation}
\subsubsection{Overall Comparision.}
The performance results of evaluating each model for graph classification and regression tasks are presented in Table \ref{table-expriemnt}. As we can see, our model significantly outperforms all the baselines on both types of tasks. On the whole, we can observe that our proposed \model improves the performance over the best message passing baselines with 2.18\% and 10.4\% for classification and regression tasks, respectively.

Among all baseline approaches, the well-designed message passing models generally show the best performance, which indicates that the essential bond features of the molecule can provide chemical semantic information for molecular representation learning. Specifically, since DMPNN and CoMPT adopt the node-edge interactive scheme, they perform slightly better than AttentiveFP. As to geometry-based baseline models, MAT can take advantage of the geometric distance from the molecular graph and performs much better than SGCN, which directly encodes the 3D coordinates and can be easily affected by the coordinate systems. Although HMGNN and DimeNet can identify the distance and angle information, they learn the molecular embedding only based on the 3D geometric graph which might be noisy. Besides, these models are designed for quantum property prediction 
and may not be expert in modeling the larger molecules. By contrast, our \gnn and \model are capable of learning from the stable 2D graph and the informative 3D geometric structure. For contrastive learning methods, MoCL achieves better results than InfoGraph, showing the significance of domain knowledge for developing the contrastive strategy \liadd{without changing the chemical semantics\hide{ of molecules}}. However, the failure of leveraging the critical geometric information limits their ability to model the molecule without labels, while our method can capture geometry-aware structural information by contrasting the 2D-3D geometric views. Therefore, \model is much effective for molecular representation learning and can accurately predict each targeted property.

\begin{figure}
  \centering
  \subfigure[Results for Classification]{
    \label{exp-ablation-cls} 
    \includegraphics[width=0.48\columnwidth]{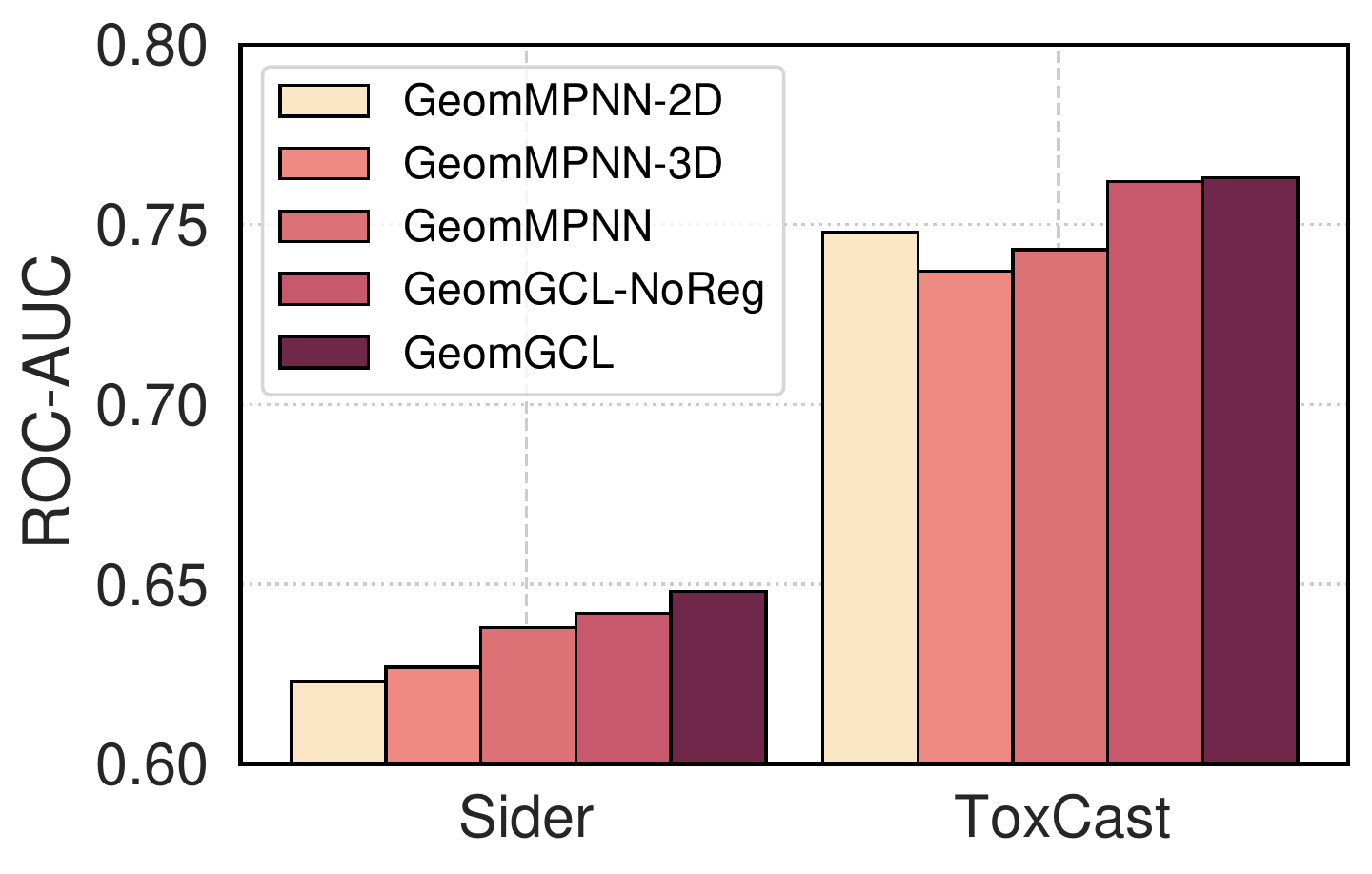}}
      \subfigure[Results for Regression]{
    \label{exp-ablation-reg} 
    \includegraphics[width=0.48\columnwidth]{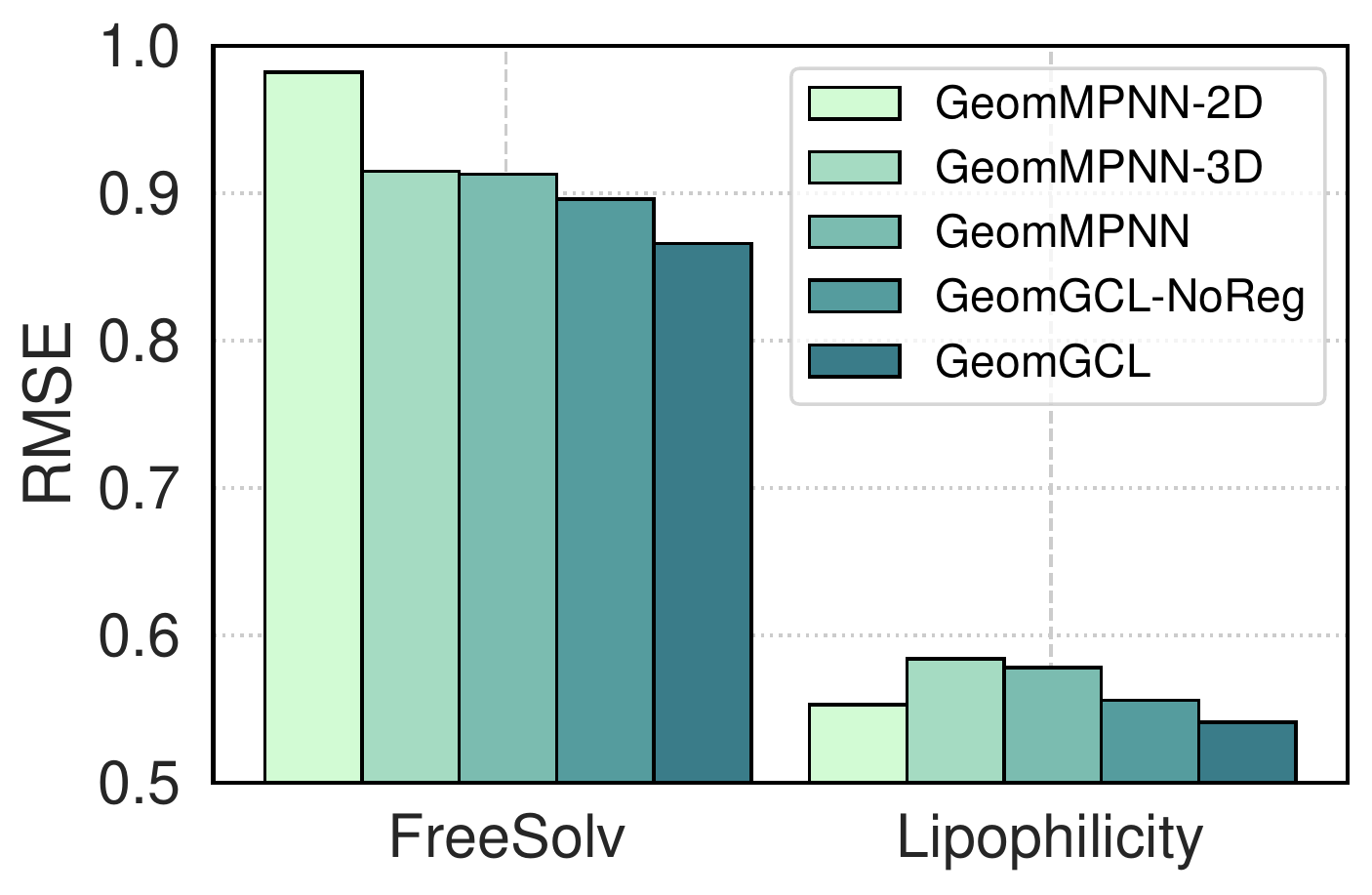}}
  \vspace{-2mm}
  \caption{Evaluation of \model with its variants.}
  \vspace{-4mm}
  \label{exp-ablation} 
\end{figure}

\subsubsection{Ablation Study.}

To further investigate the \hide{geometric }factors that influence the performance of the proposed \model framework, we conduct the ablation study on four benchmark datasets for classification and regression tasks with designing different variants of \model.
\begin{itemize}
    \item \B{\gnn-2D} only preserves the single-channel 2D geometric message passing layers.
    \item \B{\gnn-3D} only preserves the single-channel 3D geometric message passing layers.
    \item \B{\gnn} uses the dual-channel geometric message passing layers without contrastive learning.
    \item \B{\model-NoReg} removes the spatial regularizer $\mathcal{L}^{s}$.
\end{itemize}

As shown in Figure \ref{exp-ablation}, \model achieves the best performance among all architectures, proving the necessity of learning the 2D-3D geometric structures contrastively and synergistically. To be specific, we can find that learning representations from a single 2D or 3D view can not always perform better than the other view across different datasets, which supports our hypothesis that only a geometric view of molecular graph is not sufficient. Therefore, decoupling 2D or 3D geometric message passing layers from \model yields a significant drop in performance. Additionally, the use of spatial regularizer when training the model can help \model to discriminate the relative \hide{connection}correlations of different angle domains and then contributes to the performance improvements. What's more, \gnn significantly performs worse than \model, which confirms that our geometric contrastive learning scheme is beneficial for molecular representation learning.

\begin{figure}[t]
\centering
\includegraphics[width=0.7\columnwidth]{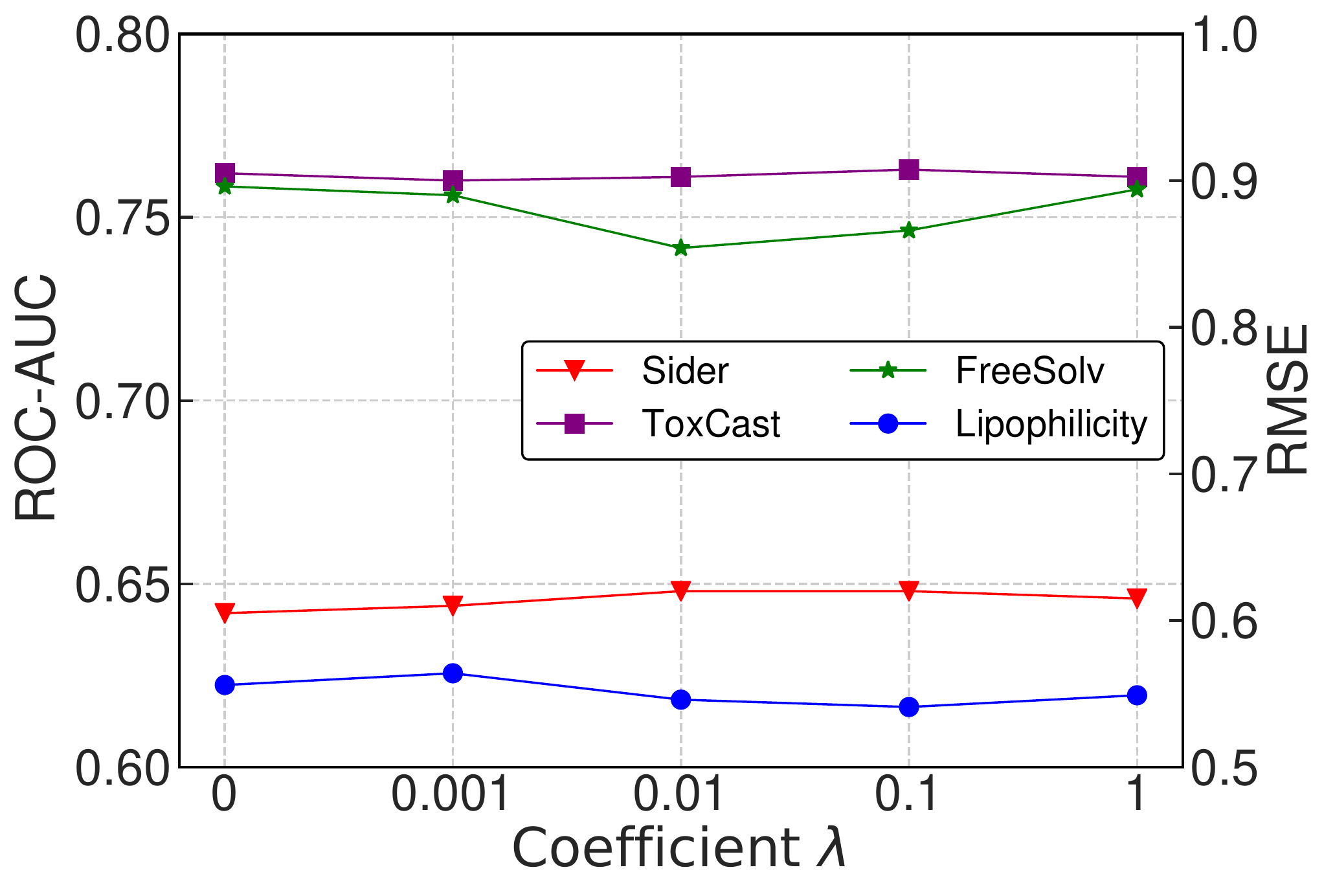}
\vspace{-1mm}
\caption{Analysis for the balancing parameter of $\mathcal{L}^{s}$.}
\label{exp-parameter}
\vspace{-4mm}
\end{figure}
\subsubsection{Parameter Analysis.}
Finally, we analyze the performance variation for \model by varying the coefficient $\lambda$ to look deeper into the impact of the spatial regularizer. As depicted in Figure \ref{exp-parameter},  we observe that the performance first tends to get better with incorporating more 3D angle domain information while training the model, and then begins to drop off slightly. The appropriate trade-off weight can assist the model in identifying the geometric factors and enhancing the representation learning. Overall, the performance of \model is stable and always better than baseline methods.
\section{Conclusion}
In this paper, we propose a novel geometric graph contrastive learning framework named \model for molecular \hide{property prediction}representation learning, which builds the bridge between the geometric structure learning and the graph contrastive learning. Along this line, we design the dual-channel geometric message passing neural networks to sufficiently capture the distance and angle information under both 2D and 3D views. Then the appropriate geometry-based contrastive learning strategy is proposed to enhance the molecular representation learning with the spatial regularizer. The experimental results on the downstream property prediction tasks demonstrate the effectiveness of the proposed \model.


\bibliography{aaai22.bib}

\end{document}